\documentclass[conference]{IEEEtran}
\IEEEoverridecommandlockouts
\usepackage{cite}
\usepackage{amsmath,amssymb,amsfonts}
\usepackage{algorithmic}
\usepackage{graphicx}
\usepackage{textcomp}
\usepackage{xcolor}
\usepackage{subfigure}
\usepackage{caption}
\usepackage{moreverb,url}
\usepackage{float}
\usepackage{gensymb}
\usepackage{booktabs}
\usepackage{subfig}
\usepackage{ragged2e}
\def\BibTeX{{\rm B\kern-.05em{\sc i\kern-.025em b}\kern-.08em
    T\kern-.1667em\lower.7ex\hbox{E}\kern-.125emX}}

\begin{document}

\title{A Miniature Biological Eagle-Eye Vision System for Small Target Detection\\
\thanks{This work was supported by the National Key Research and Development Program of China under Grant 2019YFB1703603, the National Natural Science Foundation of China under Grant 61803025 and Grant 62073031, the Interdisciplinary Research Project for Young Teachers of USTB (Fundamental Research Funds for the Central Universities) under Grant FRF-IDRY-19-010 and the Fundamental Research Funds for the China Central Universities of USTB under Grant FRF-TP-19-001C2. Corresponding author is W. He.}
}

\author{\IEEEauthorblockN{1\textsuperscript{st} Shutai Wang}
\IEEEauthorblockA{\textit{School of Automation and Electrical Engineering}\\
\textit{and Institute of Artificial Intelligence} \\
\textit{University of Science and Technology Beijing}\\
Beijing, China \\
wst19980122@126.com}
\and
\IEEEauthorblockN{2\textsuperscript{nd} Qiang Fu}
\IEEEauthorblockA{\textit{School of Automation and Electrical Engineering}\\
	\textit{and Institute of Artificial Intelligence} \\
	\textit{University of Science and Technology Beijing}\\
	Beijing, China \\
	fuqiang@ustb.edu.cn}
\and
\IEEEauthorblockN{3\textsuperscript{rd} Yinhao Hu}
\IEEEauthorblockA{\textit{School of Automation and Electrical Engineering}\\
	\textit{and Institute of Artificial Intelligence} \\
	\textit{University of Science and Technology Beijing}\\
	Beijing, China \\
	huyinhao2021@163.com}
\and
\IEEEauthorblockN{4\textsuperscript{th} Chunhua Zhang}
\IEEEauthorblockA{\textit{Automation Research Institute Co.} \\
\textit{Ltd. of China South Industries Group Corporation}\\
Mianyang, China \\
swaizhang@sina.com}
\and
\IEEEauthorblockN{5\textsuperscript{th} Wei He}
\IEEEauthorblockA{\textit{School of Automation and Electrical Engineering}\\
	\textit{and Institute of Artificial Intelligence} \\
	\textit{University of Science and Technology Beijing}\\
	Beijing, China \\
	weihe@ieee.org }

}

\maketitle

\begin{abstract}
Small target detection is known to be a challenging problem. Inspired by the structural characteristics and physiological mechanism of eagle-eye, a miniature vision system is designed for small target detection in this paper. First, a hardware platform is established, which consists of a pan-tilt, a short-focus camera and a long-focus camera. Then, based on the visual attention mechanism of eagle-eye, the cameras with different focal lengths are controlled cooperatively to achieve small target detection. Experimental results show that the designed biological eagle-eye vision system can accurately detect small targets, which has a strong adaptive ability. 
\end{abstract}


\section{Introduction}
\begin{figure}[H]
	\centering
	\subfigure[Eagle-eye]{\includegraphics[height=3cm,width=4cm]{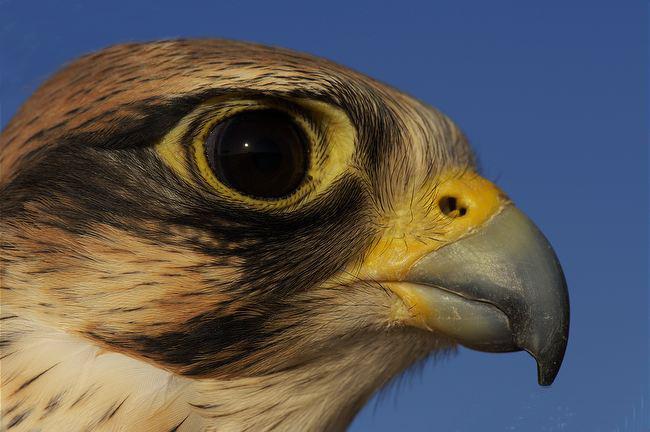}}
	\subfigure[Foveal section view]{\includegraphics[height=3cm,width=3cm]{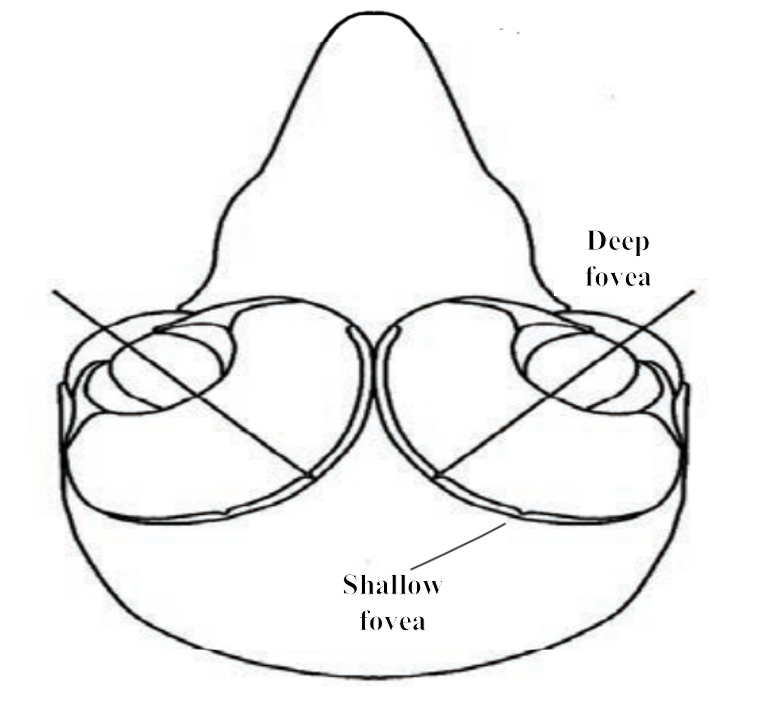}}
	\caption{Egle-eye and foveal section view}
	\label{1}
\end{figure}

Object detection is a significant research problem in the field of machine vision. At present, many researches are carried out on the situation that the distance to the target is close, and the target occupies a large proportion of the field of view. However, in aerial photography, the target in the field of view is usually small in size \cite{2013A}, low in resolution, high in noise, and very few features can be extracted \cite{2013Small}. Although target detection algorithms based on deep learning can improve performance by increasing the types and numbers of small target samples in the training set, the adaptability and accuracy of many algorithms are difficult to guarantee when dealing with small target detection. In vision tasks such as target searching \cite{2014SS} and target positioning \cite{2017Vision}, traditional vision systems have significant limitations in the efficiency and accuracy of image processing . However, the thorny issues are simple for the creatures \cite{2017Progresses}. 
Among all creatures, eagle-eye is the best in visual acuity \cite{8378050}, as shown in Fig. \ref{1}(a). Even while flying at an altitude of more than a kilometer, eagle-eye can also see the ground prey clearly \cite{2003The}, which is called clairvoyance. 

By studying the structure of eagle-eye, the foveal section view of eagles is unique \cite{2000The}. As shown in Fig. \ref{1}(b), eagle-eye has a special fovea structure to extract the image features such as intensity, color and orientation. The double-fovea structure enables the eagle to have a wide field of view \cite{2012Visual} and local high resolution \cite{1985Spatial} at the same time. 
Then, applying visual attention mechanism of the eagle-eye to the image processing process can improve the efficiency and accuracy of target searching \cite{2007Avian}. It can be seen that the structural characteristics \cite{2007Visual} and physiological mechanism of the eagle-eye are very suitable for small target detection. Therefore, the research of the biological eagle-eye vision system is of great significance to small target detection. 

At present, many researches on the structural characteristics and physiological mechanism of eagle-eye have been carried out for decades. 
Lin et al. from the National University of Singapore used the features of eagle-eye to obtain different resolutions and developed a composite vision system for multiple depth-of-field vehicle tracking and speed detection \cite{2014Biologically}. The team of Professor Duan Haibin from Beihang University has designed a visual measurement method based on the physiological mechanism of eagle-eye, which is used for the precise docking of unmanned aerial vehicles during autonomous aerial refueling \cite{2014Biological}.

Because the information processing mechanism of eagle-eye is very consistent with the visual perception requirements of flapping-wing air vehicles \cite{2017Research}\cite{Modeling2021}, the biological eagle-eye vision system is particularly suitable for flapping-wing air vehicles \cite{2021Obstacle} to perform visual tasks. 
At present, the size and weight of these vision systems are too large for flapping-wing aerial vehicles \cite{2019FW}\cite{Efficient2020}. Inspired by the structural characteristics and physiological mechanism of eagle-eye, a miniature eagle-eye vision system for small target detection is proposed in this paper. The case where the target occupies no more than 0.5\% of the image is defined as a small target in this paper. The system is divided into hardware part and software part, which has the advantages of small size and light weight. The hardware simulates the double fovea structure of eagle-eye.
A short-focus camera was selected to simulate the deep fovea of the eagle eye to obtain a large field of view, and a long-focus camera was selected to simulate the shallow fovea to achieve the function of local high resolution. And two cameras with different focal lengths simulate the zoom function. In the software part, the visual attention mechanism \cite{1998A} of eagle eye is applied to the image processing process to complete the searching target location. This technique 
can not only increase the speed of image processing, but also reduce the burden on the computer. 

The main contributions of the paper are as follows: 1) Inspired by the double-fovea structure of eagle-eye, a miniature biological eagle-eye vision system is proposed for small target detection, which has the advantages of small size and light weight; 2) The application of visual attention mechanism in image processing and analysis improves the efficiency and accuracy of target searching and target localization; 3) A cooperative control method for long and short-focus cameras is designed, which can make the target always appear near the center of the field of view of the long-focus camera to obtain more details. 


\section{Biological Eagle-Eye Vision Imaging Devie}
\subsection{Mechanical Structure Design}
High resolution and wide field of view are equally important in visual tasks. Combining the structural and functional characteristics of the eagle-eye, a biological eagle-eye visual imaging device as shown in Fig. 2 is designed to simulate a wide field of view of the eagle-eye, local high resolution and other characteristics. The device adopts a dual-camera two-degree-of-freedom structure. The short-focus camera is fixed 
\begin{figure}[H]
 	\centering
 	\includegraphics[width=0.55\linewidth, height=0.14\textheight]{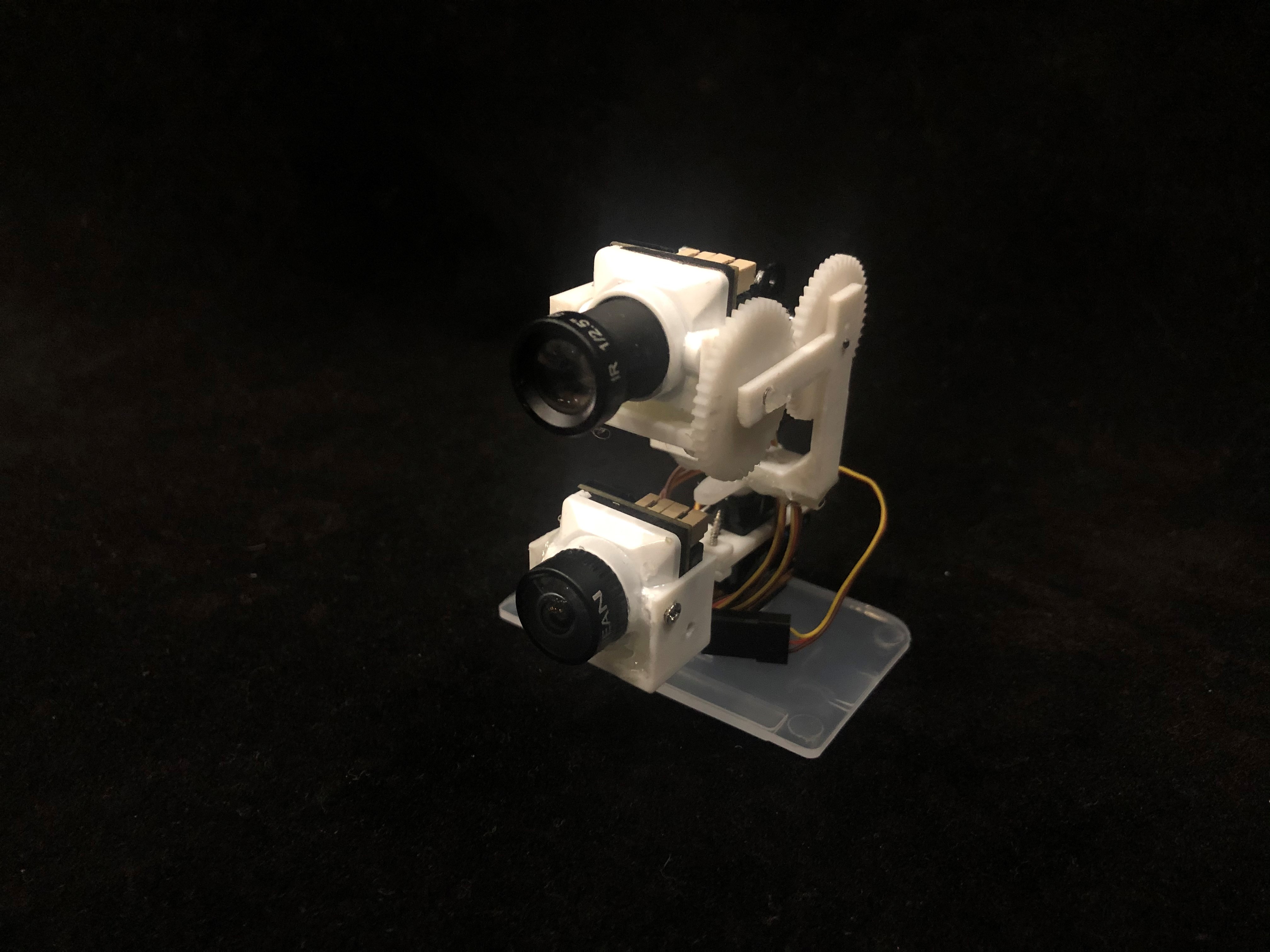}
 	\caption{The biological eagle-eye visual imaging device}
 	\label{3}
\end{figure}

\noindent to simulate the function of a large field of view. The long-focus camera can rotate two degrees of freedom. through the coperative control of the long and short focus cameras, the target always appears near the center of the field of view of the long-focus camera to obtain more details. 

\begin{figure}[H]
	\centering
	\includegraphics[width=1\linewidth, height=0.18\textheight]{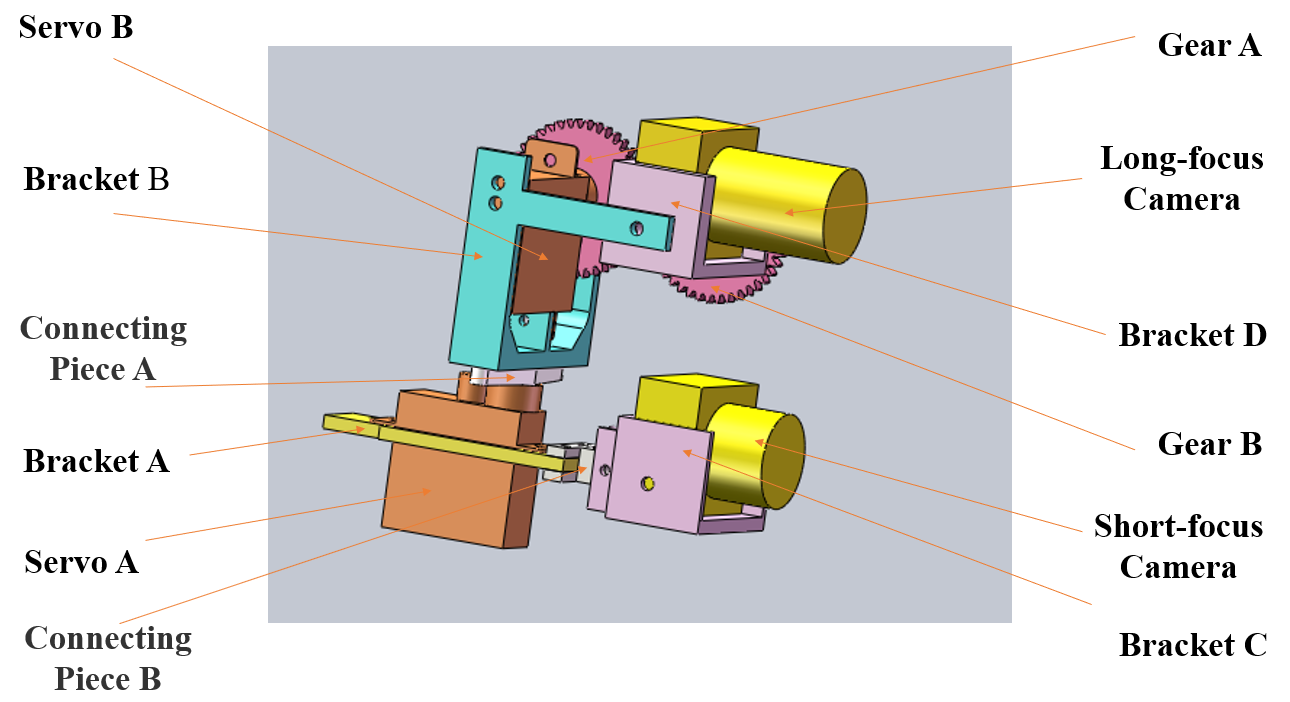}
	\caption{Structure of the biological eagle-eye vision imaging device}
	\label{4}
\end{figure}
Fig. \ref{4} shows the structure of the device, including a short-focus camera, a long-focus camera, two servos, two gears, four brackets, and two connecting pieces. Among them, gears, brackets and connecting pieces are all made by 3D printing. The bracket A is fixed with the servo A, and is connected with the short-focus camera. The bracket B is connected with the servo A through connecting piece A to fix the servo B. The servo B controls the long-focus camera rolling. The gear A is connected to the output shaft of the servo B. And the gear B is engaged with the gear A and fixed with the bracket D.

\subsection{Image acquisition module}
\begin{figure}[H]
	\centering
	\includegraphics[height=2.5cm,width=3.7cm]{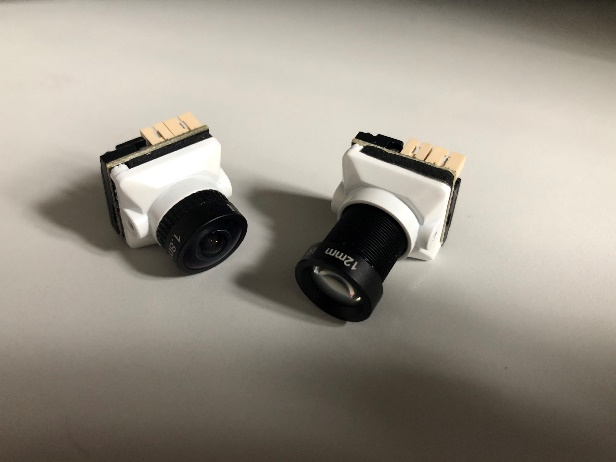}
	\caption{Short-focus camera and long-focus camera}
	\label{5}
\end{figure}

Fig. \ref{5} shows two cameras, which have different focal lengths, each weighing about 10g. The focal lengths of the two cameras are 2.1mm and 12mm respectively. 

\begin{figure}[H]
	\centering
	\subfigure[Image transmission module]{\includegraphics[height=2.5cm,width=3.2cm]{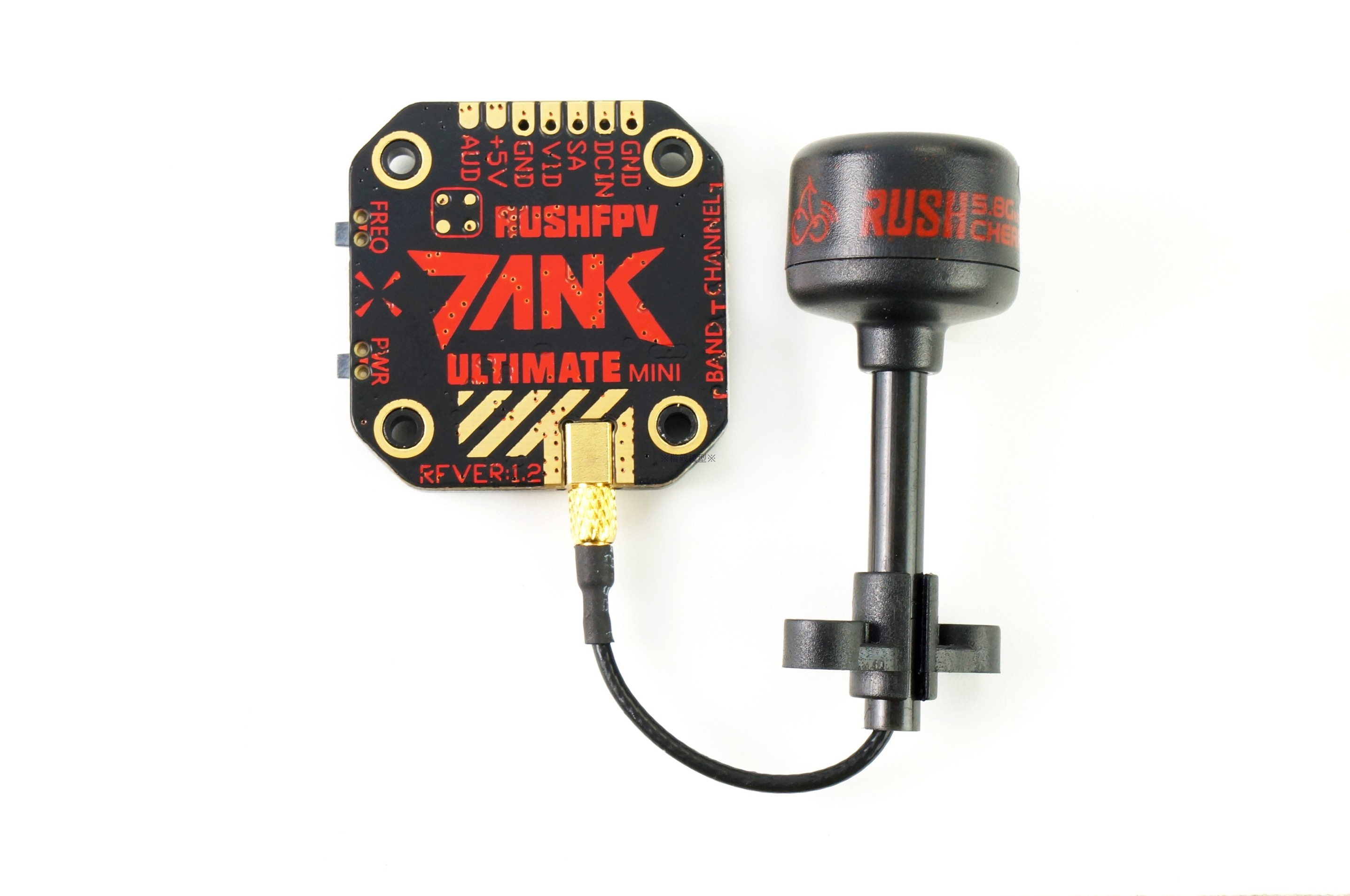}}
	\subfigure[Image receiving module]{\includegraphics[height=2.5cm,width=3.2cm]{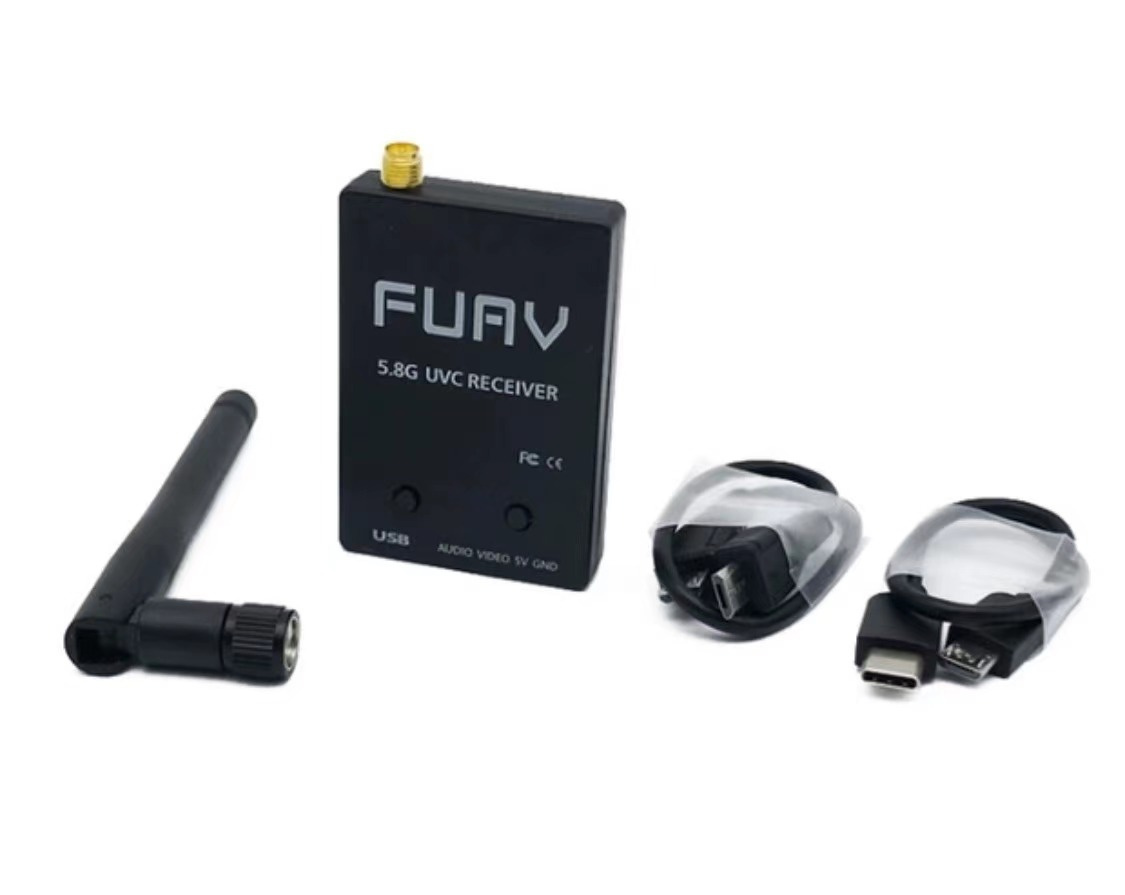}}
	\caption{ Analog image transmission module }
	\label{7}
\end{figure}

The camera is equipped with an image transmission module to transmit the image to the ground station as shown in Fig. \ref{7}(a). 
The image receiving module cooperates with the ground station to receive image information, which is given in Fig. \ref{7}(b).

The ground station is a laptop, whose configuration is given in Table \ref{t1}.
\begin{table}[h]
	\small\sf\centering
	\caption{Configuration of the ground station\label{t1}}
	\begin{tabular}{ccc}
		\toprule
		Option&Detail\\
		\midrule
		Operating system& Windows 10\\
		CPU &R7-4800H\\
		GPU &RTX-2060\\
		RAM &16G\\
		\bottomrule
	\end{tabular}\\[10pt]
\end{table}

\subsection{Control module}
\begin{figure}[H]
	\centering
	\subfigure[Control module]{\includegraphics[height=2.5cm,width=3.2cm]{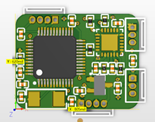}}
	\subfigure[Servo]{\includegraphics[height=2.5cm,width=3.2cm]{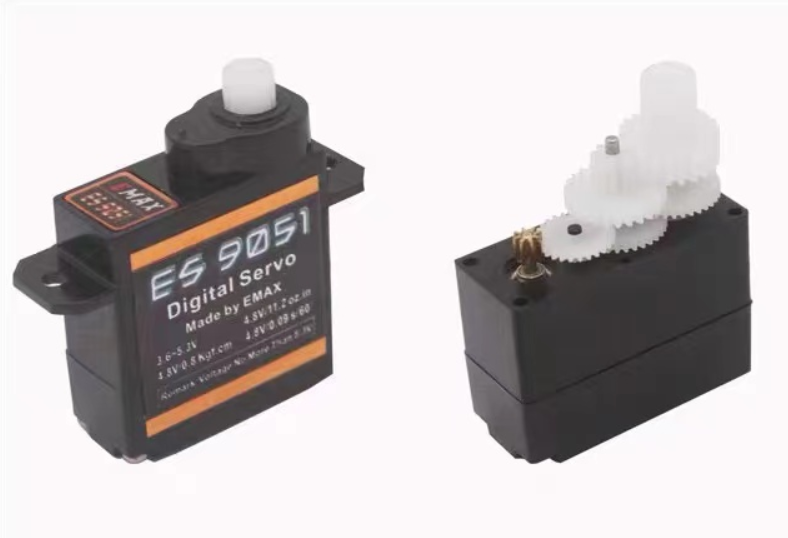}}
	\caption{Control module and servo}
	\label{8}
\end{figure}

As shown in Fig. \ref{8}(a), the servo is controlled by a self-designed control module. It is 20.45mm long, 15.74mm wide and weighs 6.24g. 

Because the servo itself contains a closed-loop control system, the output angle can be directly controlled, reducing the difficulty of development. The servo used in this paper is ES9051, as shown in Fig. 6(b). Its weight is 4.6g. 

\section{Algorithm for target detection}
\subsection{Visual attemtion mechnism}
Scientists have studied the predation process of eagles, 
and discovered that eagles use visual attention mechanisms to analyze a large amount of information and quickly search and locate targets of interest. 
This method can not only increase the speed of image processing, but also reduce the computational burden of the computer. 
Itti proposed a data-based bottom-up \cite{1608042} visual attention model in 1998. 
This model is the most representative visual attention model. 
For an input image, the model extracts primary visual features (color, intensity, and orientation) and uses center-surround operations to generate feature maps. Then all feature maps are merged into the saliency map. 
Finally, the focus of attention is obtained according to the saliency map. The detailed procedure is described as follows.


Firstly, extract early visual features. 
Choose r, g, and b to represent the three channels of red, green and blue, and perform Gaussian down-sampling on these three channels to obtain a three-channel image of nine scales, which includes $R(\sigma)$,  $G(\sigma)$ and  $B(\sigma)$. Calculate $I=\frac{r+g+b}{3}$under nine scales to get $I(\sigma)$. Then, the color Gaussian pyramid can be constructed and calculated on nine scales, where $\sigma\in\left\{ 0..8 \right\}$. The r, g, and b channels are normalized by I to decouple the hue from the intensity. 
\begin{equation}
	R(\sigma)=r(\sigma)-\frac{g(\sigma)+b(\sigma)}{2}
\end{equation}
\begin{equation}
	G(\sigma)=g(\sigma)-\frac{r(\sigma)+b(\sigma)}{2}
\end{equation}
\begin{equation}
	B(\sigma)=b(\sigma)-\frac{r(\sigma)+g(\sigma)}{2}
\end{equation}
\begin{equation}
	Y(\sigma)=\frac{r(\sigma)+g(\sigma)}{2}-\frac{|r(\sigma)-g(\sigma)|}{2}-b(\sigma)
\end{equation}

The above four formulas represent the Gaussian pyramids of red, green, blue, and yellow. Then, the model constructs the Gabor direction pyramid $O(\sigma, \theta)$ using the Gabor filter, where  $\theta\in\left\{0\degree, 45\degree, 90\degree, 135\degree\right\}$.

Secondly, construct feature maps. The model uses center-surround method to calculate the corresponding feature map. The calculation method is as follows:
\begin{equation}
	I(c, s) = |I(c)\Theta I(s)|
\end{equation}
\begin{equation}
	RG(c, s) = |(R(c)-G(c)) \Theta (G(s)-R(s))|
\end{equation}
\begin{equation}
	BY(c, s) = |(B(c)-Y(c)) \Theta (Y(s)-B(s))|
\end{equation}
\begin{equation}
	O(c, s, \theta) = |O(c, \theta) \Theta O(s, \theta)|
\end{equation}
where $c\in\left\{ 2, 3, 4 \right\}$ and $s=c+\delta$, $\delta\in\left\{3, 4 \right\}$.$\Theta$ refers to the matrix subtraction after adjusting the two images to the same size, $I(c, s)$ represents a brightness feature map, $RG(c, s)$ and $BY(c, s)$ represent a color feature map, and $O(c, s, \theta)$ represents a directional feature map.

Finally, construct saliency maps. In the absence of a top-down supervision mechanism, the model proposes a feature map normalization operator $N(\cdot)$. This operation process is based on the wit of lateral inhibition of the cerebral cortex, which can enhance the feature map with a few sharp values and suppress the feature map with many sharp values. The operation method is as follows:
\begin{equation}
	\overline {\rm{I}}  = \begin{array}{*{20}{c}}
		4\\
		\oplus \\
		{c = 2}
	\end{array}\begin{array}{*{20}{c}}
		{c + 4}\\
		\oplus \\
		{c + 2}
	\end{array}{\rm{N}}(I(c,s))
\end{equation}
\begin{equation}
	\overline C  = \begin{array}{*{20}{c}}
		4\\
		\oplus \\
		{c = 2}
	\end{array}\begin{array}{*{20}{c}}
		{c + 4}\\
		\oplus \\
		{s = c + 3}
	\end{array}{\rm{[N}}(RG(c,s)) + N(BY(c,s))]\
\end{equation}
\begin{equation}
	\begin{array}{l}
		\overline O  = \\
		\sum\limits_{\theta  \in \left\{ {0{\rm{^\circ ,}}45{\rm{^\circ ,90^\circ ,135^\circ }}} \right\}} {N\left[ {\begin{array}{*{20}{c}}
					4\\
					\oplus \\
					{c = 2}
				\end{array}\begin{array}{*{20}{c}}
					{c + 4}\\
					\oplus \\
					{s = c + 3}
				\end{array}{\rm{N}}(O(c,s,\theta ))} \right]} 
	\end{array}
\end{equation}

The above $\bigoplus$ refers to the operation of adding multiple images to the same size so that the brightness, color and direction saliency map is obtained, and the final saliency map S is as follows:
\begin{equation}
	S=\frac{N(I)+N(C)+N(O)}{3}
\end{equation}

\subsection{Cooperative control for long and short focus cameras}
\begin{figure}[H]
	\centering
	\includegraphics[width=1\linewidth, height=0.19\textheight]{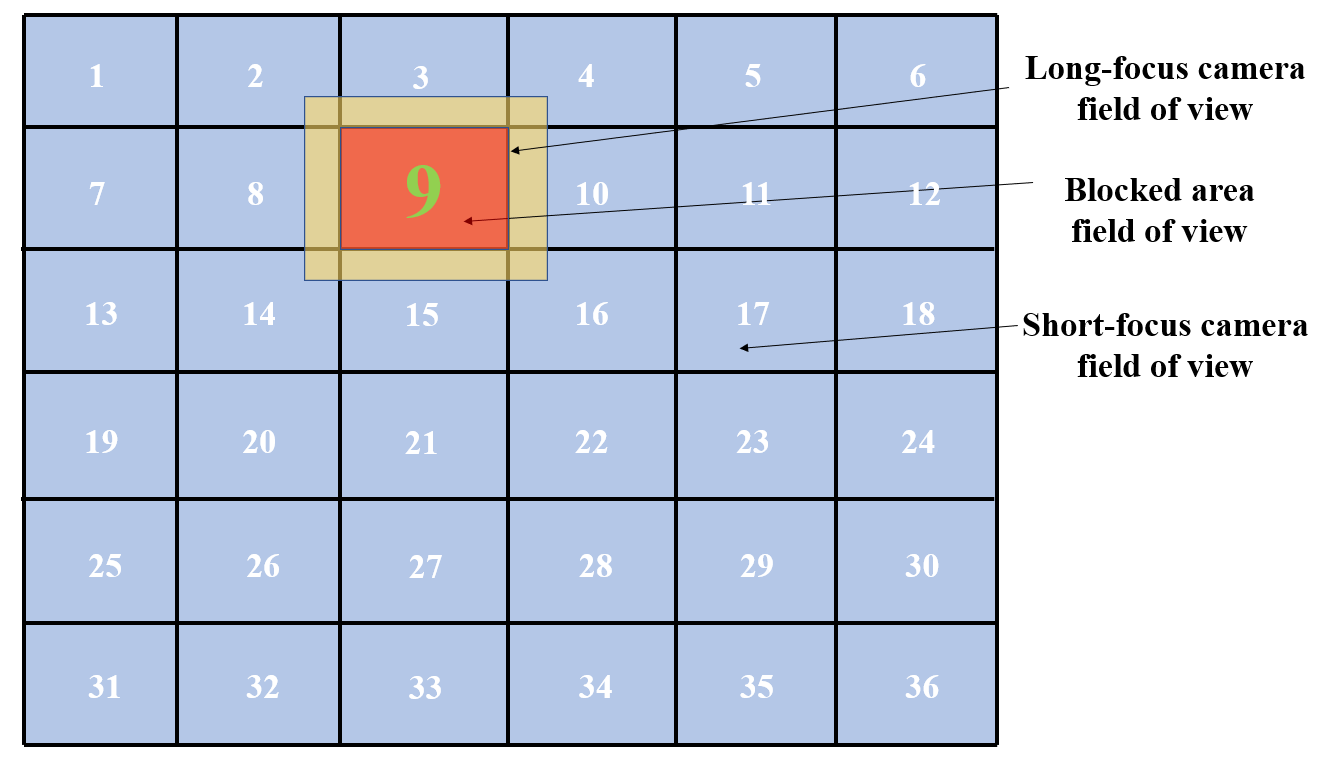}
	\caption{Field of view diagram}
	\label{10}
\end{figure}

In order to make the target always appear near the center of the field of view of the long-focus camera, the field of view of the short-focus camera is divided into blocks. The field of view of a short-focus camera is about 5 times that of a long-focus camera. The horizontal field of view and vertical field of view are divided into 6 parts to prevent target loss. Therefore, the field of view of short-focus camera is divided into 36 small areas, as shown in Fig. \ref{10}.   

The long-focus camera is located 36.88mm above the short-focus camera. Through continuous debugging, the long-focus camera aims at the center of the short-focus camera, and this is determined as the initial position. The target area is determine when the target appears. Then by controling the servos to drive the long-focus camera to the center of each area, the target can appear in the field of view of the long-focus camera at this time.

\section{Experimental results and analysis}
In this experiment, a human was selected as a reference to test the performance of the biological eagle-eye vision imaging device. The size of the target occupies approximately 0.4\% of the image. 
  
\subsection{Target searching of the short-focus camera}
\begin{figure}[H]
	\centering
	\subfigure[Original image]{\includegraphics[height=3cm,width=4cm]{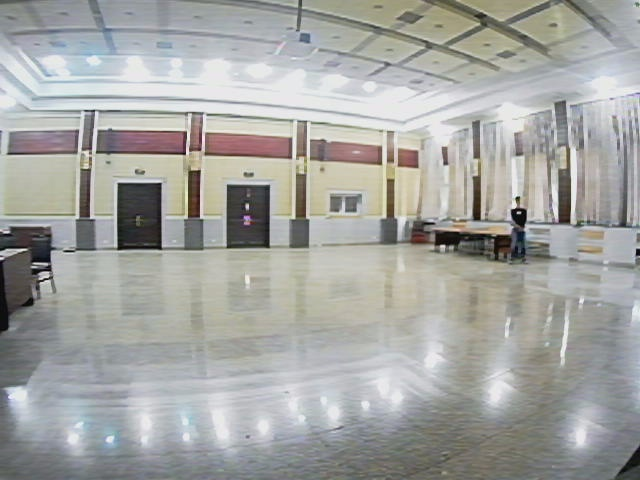}}
	\subfigure[Saliency map]{\includegraphics[height=3cm,width=4cm]{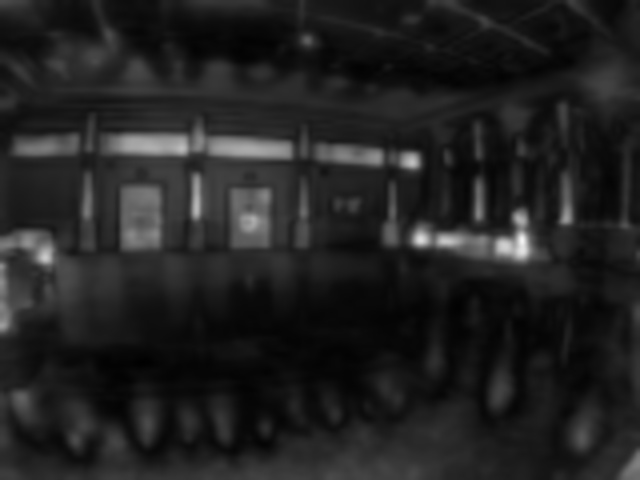}}
	\caption{Short-focus camera field of view and saliency map}
	\label{11}
\end{figure}

The Itti visual attention model is used to obtain the saliency map under the wide field of view. Fig. \ref{11} shows the field of view of the short-focus camera and the corresponding saliency map. Saliency detection can be of maximum value with limited resources.

In the saliency map, the pixel with the largest gray value is the brightest point, which represents a potential target. Through traversing all pixel points, the algorithm finds the saliency point, marks it, and outputs its gray value and coordinate value. Then Gaussian blur is applied for preprocessing, 
\begin{figure}[H]
	\centering
	\subfigure[Saliency point map]{\includegraphics[height=3cm,width=4cm]{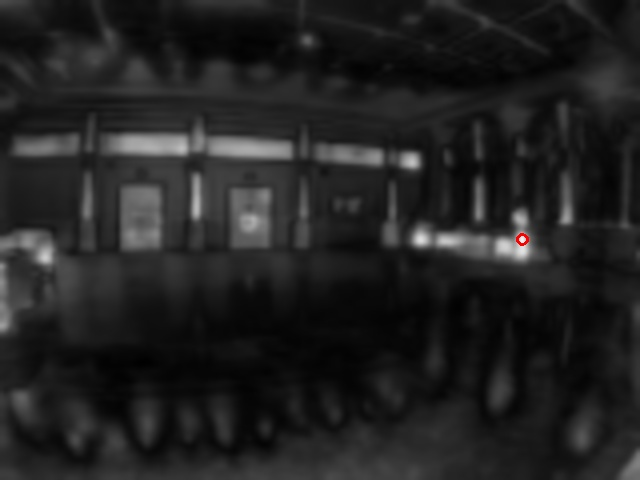}}
	\subfigure[Saliency area map]{\includegraphics[height=3cm,width=4cm]{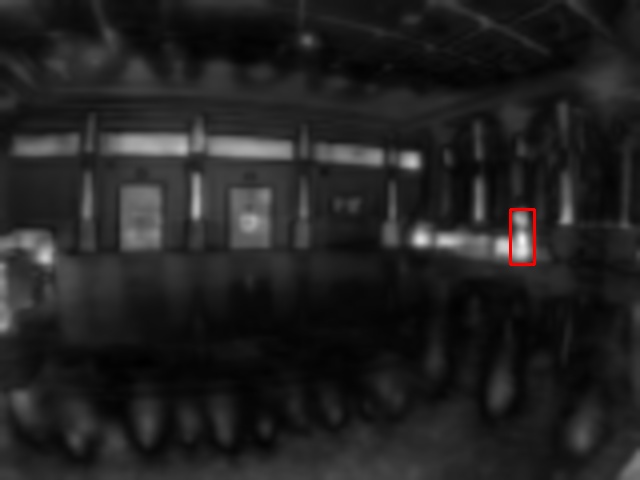}}
	\caption{Saliency point and area map}
	\label{12}
\end{figure} 
\noindent from finding a point to finding a region. The experimental result for a sample image is shown in Fig. \ref{12}.

Coordinate: (522,239), Gray value: 243


It is easy to find that the centroid coordinate of the target is approximately (530,229) by hand. By comparison, the experimental results are close to the actual values.

\subsection{Target capture of the long-focus camera}
\begin{figure}[htbp]
	\centering
	\subfigure[Target in short-focus camera]{\includegraphics[height=3cm,width=4cm]{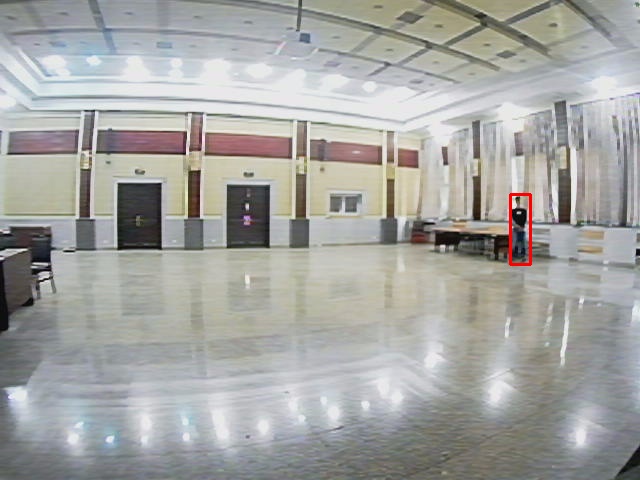}}
	\subfigure[Target in long-focus camera]{\includegraphics[height=3cm,width=4cm]{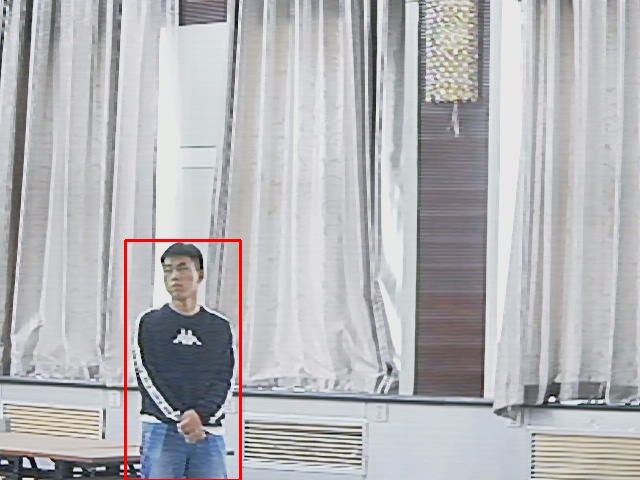}}
	\caption{Target in cameras with different focal lengths}
	\label{13}
\end{figure}

The ground station converts the coordinate of saliency point into a PWM signal and sends it to the control module. The control module controls the steering servos to drive the long-focus camera to capture the target and obtain more details. Fig. \ref{13}(a) and (b) show the target at short and long focus cameras respectively.

Through comparison, it is found that the size of the target accounts for 0.45\% of the image in the short-focus camera field of view, and 8.8\% of the image in the long-focus camera field of view. It can be seen from the experimental results that the device proposed in this paper can accurately complete the target searching, and complete the coorperative control of the long and short focus cameras, obtain more target details, and greatly improve the efficiency and accuracy of small target detection. To ensure the persuasiveness of the experiment, we conducted several sets of experiments from different angles, and the experimental results both reached the expected requirements, which proved the feasibility of the experiment.


\section{Conclusion}
This paper proposed a miniature bioligical eagle-eye vision system based on the structural characteristics and physiological mechanism of the eagle-eye, which both has a wide field of view and local high resolution. The system includes a biological eagle-eye vision imaging device and software algorithms. In addition, the imaging device was used for experiments, and the experimental results proved the feasibility of the designed biological eagle-eye vision system. 

The biological eagle-eye vision imaging device is small in size, light in weight and simple in structure. In future work, it can be used as a vision system for micro-robots such as flapping-wing air vehicles to achieve target detection, target tracking and other tasks. 

\bibliographystyle{IEEEtran}
\bibliography{ref}

%

\end{document}